\documentclass[letterpaper]{article} 
\usepackage{aaai23}
\usepackage{balance}
\usepackage{float}
\usepackage{stfloats}
\usepackage{multirow}
\usepackage{amsfonts}
\usepackage{amsmath}
\usepackage{times}  
\usepackage{helvet}  
\usepackage{courier}  
\usepackage[hyphens]{url}  
\usepackage{graphicx} 
\usepackage{natbib}  
\usepackage{caption} 
\usepackage{algorithm}
\usepackage{algorithmic}

\usepackage{newfloat}
\usepackage{listings}
\DeclareCaptionStyle{ruled}{labelfont=normalfont,labelsep=colon,strut=off} 
\floatstyle{ruled}
\newfloat{listing}{tb}{lst}{}
\floatname{listing}{Listing}
\graphicspath{{image/}}
\title{Search for or Navigate to? Dual Adaptive Thinking for Object Navigation}
\author{
    Ronghao Dang\textsuperscript{\rm 1},
    Liuyi Wang\textsuperscript{\rm 1},
    Zongtao He\textsuperscript{\rm 1},
    Shuai Su\textsuperscript{\rm 1},
    Chengju Liu\textsuperscript{\rm 1 2}\thanks{Corresponding author},
    Qijun Chen\textsuperscript{\rm 1}
}
\affiliations{
    \textsuperscript{\rm 1}Department of Control Science and Engineering, Tongji University, Shanghai 201804, China\\
    \textsuperscript{\rm 2}Tongji Artificial Intelligence (Suzhou) Research Institute, Suzhou 215000, China

    \{dangronghao, wly, xingchen327, sushuai, liuchengju, qjchen\}@tongji.edu.cn 
}

\usepackage{bibentry}

\begin{document}

\maketitle

\begin{abstract}
“Search for" or “Navigate to"? When we find a specific object in an unknown environment, the two choices always arise in our subconscious mind.  Before we see the target, we search for the target based on prior experience.  After we have located the target, we remember the target location and navigate to this location. However, recent object navigation methods almost only consider using object association to enhance the “search for” phase while neglect the importance of the “navigate to” phase. Therefore, this paper proposes a dual adaptive thinking (DAT) method that flexibly adjusts the thinking strategies in different navigation stages. Dual thinking includes both search thinking according to the object association ability and navigation thinking according to the target location ability. To make the navigation thinking more effective, we design a target-oriented memory graph (TOMG) that stores historical target information and a target-aware multi-scale aggregator (TAMSA) that encodes the relative position of the target. We assess our methods on the AI2-Thor dataset. Compared with state-of-the-art (SOTA) methods, our approach achieves 10.8\%, 21.5\% and 15.7\% increases in the success rate (SR), success weighted by path length (SPL) and success weighted by navigation efficiency (SNE), respectively. 
\end{abstract}

\section{Introduction}
Object navigation \cite{moghaddam2022foresi,li2022reve} is a challenging task that requires an agent to find a target object in an unknown environment with first-person visual observations. Due to the limited field of view, the information that guides the agent navigation process is insufficient. Therefore, some researchers recently introduced scene prior knowledge into end-to-end navigation networks. These methods have been applied to address various issues, including the use of object associations \cite{yang2018visual}, object attention bias \cite{dang2022unbiased}, and the lack of universal knowledge \cite{gao2021room}. However, these methods improve the efficiency of only the ``search for'' phase (start$\rightarrow$first seeing the target) while neglecting the ``navigate to'' phase (first seeing the target$\rightarrow$end). 
Our experiments (Table~\ref{tab:single-dual-human} in supplementary material) show that for the current SOTA end-to-end methods, the “navigate to” steps accounts for 60\% of the whole path, while only 40\% for humans; the success rate after seeing the target is only 80\%, while humans can reach 100\%. 

Some modular approaches \cite{chaplot2020object,ramakrishnan2022poni} model the environment  by using top-down semantic maps. With the help of the detailed semantic maps, the object navigation task can be decoupled into two separate training subtasks: predicting the subtarget point and navigating to the subtarget point, thus optimizing the agent navigation ability after seeing the target. However, these methods depend strongly on semantic maps, which are hypersensitive to sensory noise and scene changes. Furthermore, the generation of high-quality semantic maps requires a large amount of computational resources. 

\begin{figure}[t]
\centering
\includegraphics[width=0.95\columnwidth]{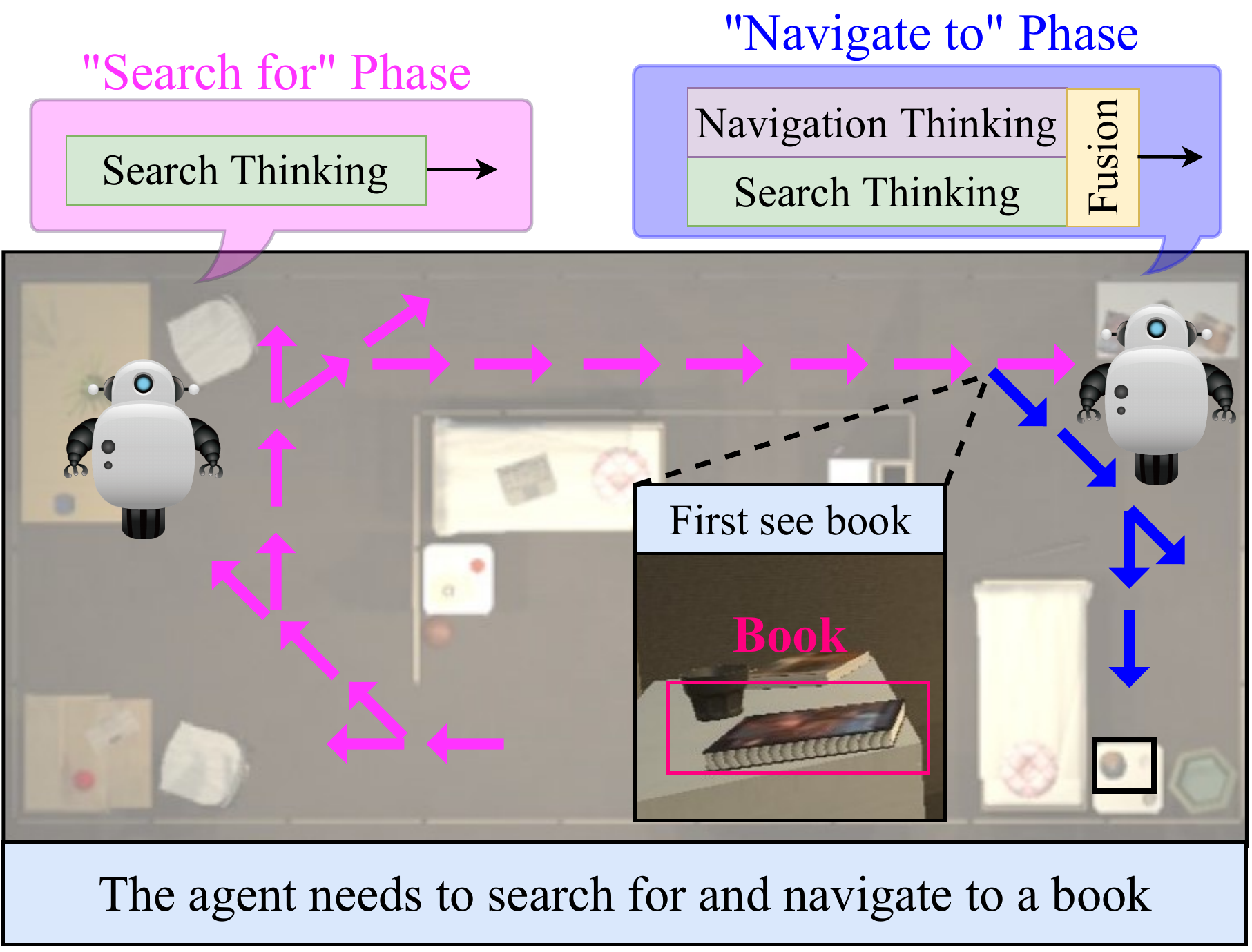} 
\caption{We divide the agent's navigation process into two phases: “search for" (pink) and “navigate to" (blue). During the “search for" phase, the agent uses only search thinking to search for the target. During the “navigate to" phase, navigation thinking assists the agent in quickly navigating to the target location.}
\label{first_figure}
\end{figure}

To address the above issues, we aim to integrate this task decoupling concept in modular methods into end-to-end methods. Therefore, we propose the dual adaptive thinking (DAT) method. As shown in Figure~\ref{first_figure}, the agent's thinking modes are divided into search thinking and navigation thinking. Search thinking guides the agent to quickly locate the target according to prior knowledge. Navigation thinking assists the agent in efficiently navigating to the target position after locating the target. The agent adaptively adjusts the dominance of the two thinking ways in an end-to-end network according to the navigation progress. 

Specifically, we develop different designs for the search thinking network and navigation thinking network. For the search thinking network, we adapt the directed object attention (DOA) graph method proposed in \cite{dang2022unbiased} to design object association and attention allocation strategies. For the navigation thinking network, we propose a target-oriented memory graph (TOMG) to store the simplified agent state and target orientation information. Furthermore, we design a target-aware multi-scale aggregator (TAMSA) to refine the features in the TOMG to guide the agent's navigation.

Extensive experiments on the AI2-Thor \cite{Kolve2017AI2THORAn} dataset show that our dual adaptive thinking (DAT) method not only optimizes the “navigate to” phase in the end-to-end network but also outperforms the state-of-the-art (SOTA) method \cite{dang2022unbiased} by 10.8\% and 21.5\% in the success rate (SR) and success weighted by path length (SPL). Moreover, we propose a new metric, success weighted by navigation efficiency (SNE), to assess the agent's navigation ability during the “navigate to” phase.
As a general concept, the proposed multiple thinking strategy can be applied in various other embodied artificial intelligence tasks. Our contributions can be summarized as follows:

\begin{itemize}
\item We propose a dual adaptive thinking (DAT) method that allows the agent to flexibly use different modes of thinking during navigation.
\item We carefully design a navigation thinking network with a selective memory module (TOMG) and a feature refinement module (TAMSA).
\item We demonstrate that our DAT method not only addresses inefficiencies in the “navigate to" phase but also substantially outperforms existing object navigation models.
\end{itemize}

\section{Related Works}
\subsection{Object Navigation}

Object navigation tasks require an agent to navigate to a target object in an unknown environment while considering only visual inputs. Recently, the relationships between objects have been introduced into navigation networks, allowing agents to locate targets more quickly by considering object associations. In \cite{zhang2021hierarchical}, the hierarchical object-to-zone (HOZ) graph was used to guide an agent in a coarse-to-fine manner. Moreover, researchers \cite{dang2022unbiased} have utilized the directed object attention (DOA) graph to address the object attention bias problem. Although these works allow agents to locate targets faster, they do not address the problem of how to navigate to these targets more quickly. Our dual adaptive thinking (DAT) method divides the thinking into two types: search thinking and navigation thinking, which can collaborate adaptively to make every navigation stage efficient. 


\subsection{Modular Navigation}

The modular navigation method has been proposed to solve the generalizability problem of end-to-end models in complex environments. It has been proven that using a top-down semantic map to predict distant subgoal points \cite{chaplot2020object} is feasible on the Habitat dataset. The PONI \cite{ramakrishnan2022poni} method trains two potential function networks using supervised learning to determine where to search for an unseen object. These modular methods require a large amount of computing and storage resources to generate semantic maps in real time and are sensitive to the image segmentation quality. Our method implicitly incorporates different thinking during navigation into an end-to-end network without relying on semantic maps.

\begin{figure*}[t]
\centering
\includegraphics[width=0.95\textwidth]{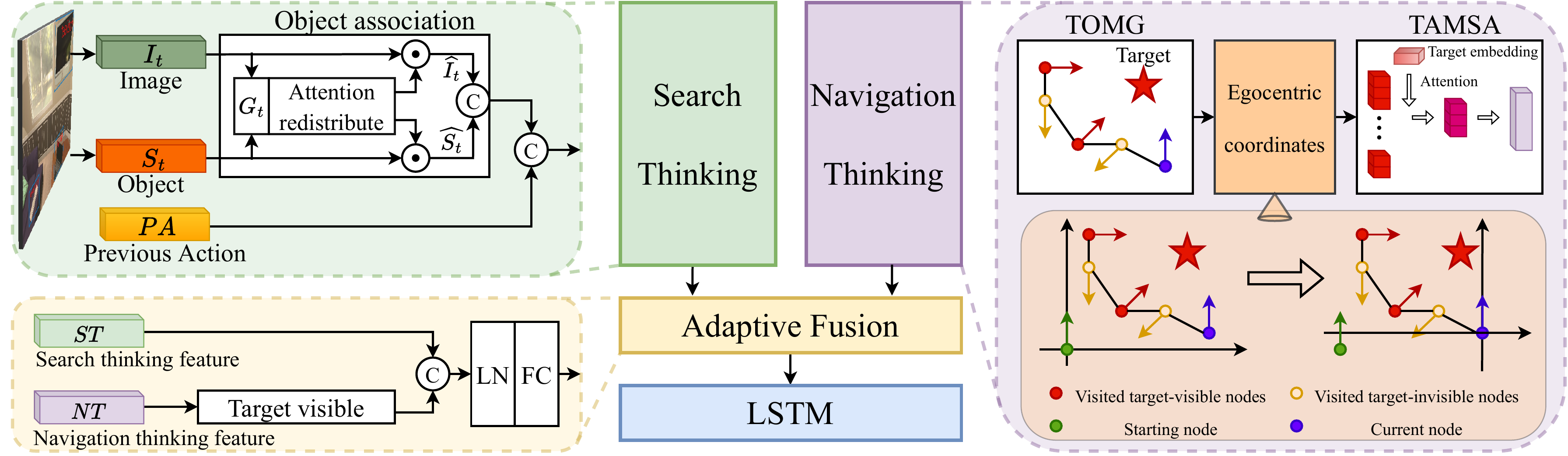} 
\caption{Model overview. TOMG: target-oriented memory graph. TAMSA: target-aware multi-scale aggregator. Our model includes three modules: search thinking, navigation thinking and adaptive fusion. In the search thinking network, we endow the model with an object association  ability according to the DOA graph method proposed in \cite{dang2022unbiased}. In the navigation thinking network, we provide the model with the ability to remember the target orientation. In the adaptive fusion network, we make the dual thinking work in harmony according to the navigation progress.}
\label{model_architecture}
\end{figure*}

\section{Necessity of Dual Thinking}
\subsection{Dual Thinking in Humans}
Embodied AI \cite{duan2022survey} is a challenging research topic that requires agents to use well-developed intuitive tasks (e.g., classification \cite{wang2019development} and detection \cite{liu2020deep}) to complete complex logical tasks (e.g., navigation \cite{zhu2022integrity} and interaction \cite{shridhar2020alfred}) in real-world environments. Humans often use more than one way of thinking when completing these complex logical tasks. For example, when we need an object, we first use associative thinking to locate the object and then use navigational thinking to reach the object location; when we answer a question about an object, we first use exploratory thinking to fully understand the object and then use reasoning and language-organized thinking to draw conclusions. Therefore, multiple thinking approaches can be introduced in end-to-end networks to develop interpretable hierarchical models that are more consistent with how humans address complex logic problems.

\subsection{Repeated Target Search Problem}
There is a phenomenon in the current methods that if the agent loses the target in view, it still needs to be searched again to locate the target. This phenomenon causes the agent to waste a considerable amount of time while re-searching for the target, potentially leading to constant loops. This problem is especially common in environments with many obstacles. A clear orientation memory for the target is the key to solve this problem. Therefore, we design a target-oriented memory graph (TOMG) and target-aware multi-scale aggregator (TAMSA) in the navigation thinking network to ensure that the agent navigates to the target efficiently without repeatedly re-searching.

\section{Dual Adaptive Thinking Network}

Our goal is to endow agents with both search and navigation thinking and to adjust their status based on the navigation process. To achieve this goal, we design three networks, as illustrated in Figure~\ref{model_architecture}: (\romannumeral1) search thinking network; (\romannumeral2) navigation thinking network; (\romannumeral3) adaptive fusion network. (\romannumeral1) and (\romannumeral2) are connected by (\romannumeral3) to form the dual adaptive thinking (DAT) network.

\subsection{Task Definition}
The agent is initialized to a random state $ s=\{x,y,\theta,\beta \}$ and random target object $p$. According to the single view RGB image $o_t$ and target $p$, the agent learns a navigation strategy $\pi(a_t|o_t, p)$, where $a_t\in A=\{MoveAhead$; $RotateLeft$; $RotateRight$;  $LookDown$;  $LookUp$; $Done\}$ and $Done$ is the output if the agent believes that it has navigated to the target location.
Ultimately, if the agent is within 1.5 m of the target object when $Done$ is output, the navigation episode is considered successful. 

\subsection{Search Thinking Network}
Search thinking aims to enable the agent to quickly capture the target with the fewest steps when the target is not in view. To use efficient object association, we adopt the unbiased directed object attention (DOA) graph method proposed in \cite{dang2022unbiased}. According to the object-target association score $G_t$ calculated by the DOA method, we redistribute the attention to the object features $S_t$ (from DETR \cite{carion2020end}) and image features $I_t$ (from ResNet18 \cite{he2016deep}) to ensure that the agent pays attention to objects and image regions that are more relevant to the target. 

In the object attention redistribution process, the object-target association score of each object $q$ is multiplied by the object features $S_t$ to generate the final object embedding $\widehat {S}_{t}$: 
\begin{equation}
 \widehat{S}_{t}^{q} = S_{t}^{q} G_{t}^{q} \;\;\;\;\;\;\;\;
 q = 1,2,\cdots, N
\end{equation}
where $\widehat{S}_{t}=\{\widehat{S}_{t}^{1},\widehat{S}_{t}^{2},\cdots ,\widehat{S}_{t}^{N}\}$, and $N$ is the number of objects. 

In the image attention redistribution process, we assign attention to image features $I_t$ according to the object semantic embeddings generated by the one-hot encodings. Initially, the semantic embeddings are weighted by $G_t \in \mathbb{R}^{N \times 1}$ to obtain the attention-aware object semantics $D$. We use $D$ as the query and $I_t$ as the key and value in the multi-head image attention to generate the final image embedding $\widehat {I}_{t}$:
\begin{small}
\begin{gather}
Q_i = D W_i^Q \;\;
K_i= I_t W_i^K \;\;
V_i= I_t W_i^V \;\;
i=1,\cdots ,NH\\
head_i=softmax(\frac {Q_{i} K_{i}^T}{\sqrt{HD}})V_i \\
\widehat {I}_{t} = Concat(head_1,\cdots , head_{NH})W^O
\end{gather}
\end{small}
where $HD$ and $NH$ denote the hidden dimensionality and number of heads in the multi-head attention. 

Finally, the attention-aware object features $\widehat {S}_{t}$ and image features $\widehat {I}_{t}$ are concatenated with the previous action embedding $PA$ to obtain the output $ST$ of the search thinking network.

\subsection{Navigation Thinking Network}
\subsubsection{Target-Oriented Memory Graph (TOMG)}
In contrast to search thinking, navigation thinking requires the ability to memorize, locate and navigate to the target. Thus, we design a target-oriented memory graph (TOMG) as the input feature $M$. As shown in Figure~\ref{model_architecture}, the TOMG is composed of the visited target-visible nodes. Each node feature $m \in \mathbb{R}^{1 \times 9}$ is concatenated by three parts: the target bounding box, the target confidence and the agent's state (position and angle). This target-oriented method of storing information about visited nodes uses $400\times$ less storage than the methods used in previous works \cite{fukushima2022object,zhu2021soon}. Since the agent cannot obtain its own absolute position and orientation in unknown environments, the stored coordinates take the starting position as the origin and the starting orientation as the coordinate axis. Target-visible nodes are filtered by a confidence threshold $cf$. Finally, to reduce the storage redundancy, only the $L$ closest target-visible nodes to the current node in the path are stored. If the number of target-visible nodes is less than $L$, the remaining nodes are filled with values of 0. 

\subsubsection{Egocentric Coordinate Transformation}
In the above section, we mentioned that the agent's position $(x_i, y_i)$ and angle $( \theta _i , \beta _i)$ are calculated relative to the starting position $(x_0, y_0)$ and angle $( \theta _0, \beta _0)$. However, as the agent navigates during each step, the decisions (e.g., rotate right) are made relative to the agent's own coordinate system.  Therefore, as shown in Figure ~\ref{model_architecture}, we convert the coordinates of each node in the TOMG to the coordinate system of the current node $(x_c, y_c, \theta _c , \beta _c)$ at each step:
\begin{equation}
\begin{aligned}
(\widetilde {x}_i, \widetilde {y}_i)&=(x_i,y_i)-(x_c,y_c)\\
(\widetilde {\theta} _i^{x}, \widetilde {\beta } _i^{x})&=sin((\theta _i,\beta _i)-(\theta  _c,\beta _c))\\
(\widetilde {\theta} _i^{y}, \widetilde {\beta } _i^{y})&=cos((\theta _i,\beta _i)-(\theta _c,\beta _c))\;\;\;i\in \Delta _M
\end{aligned}
\end{equation}
where $\Delta _M$ represents the index collection of target-visible nodes. To ensure that the angle and position coordinates have the same order of magnitude, we use $sin$ and $cos$ to normalize the angle coordinates to $[-1, 1]$. After this egocentric coordinate transformation, we obtain egocentric TOMG features $ \widetilde{M} \in \mathbb{R}^{L\times 11}$.

\subsubsection{Target-Aware Multi-Scale Aggregator (TAMSA)}
\begin{figure}[t]
\centering
\includegraphics[width=0.96\columnwidth]{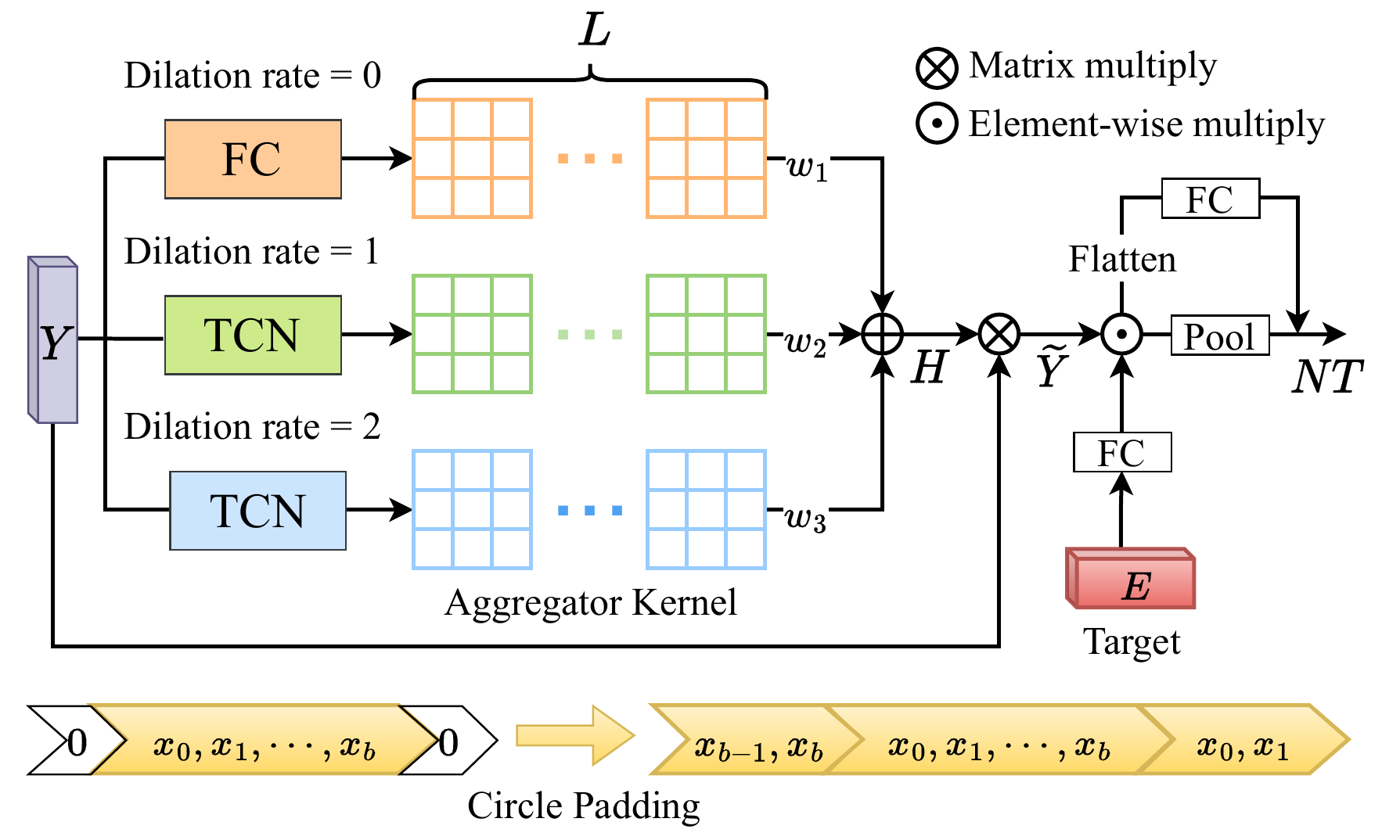} 
\caption{A detailed explanation of the target-aware multi-scale aggregator (TAMSA). We first use the multi-scale TCNs to obtain aggregator kernels that aggregate the target-oriented memory graph with $L$ nodes into graph with $\pounds$ nodes. Then, the aggregated features allocate attention to the channel dimension using the target semantics. We describe the circle padding method applied in our TCNs below the figure.}
\label{TAMSA}
\end{figure}

To encode navigation thinking into the network, we design a target-aware multi-scale aggregator (TAMSA) to aggregate the egocentric TOMG feature $ \widetilde{M}$ into an implicit representation $NT$.
In contrast to typical methods that use transformers or temporal convolutions as encoders, we devise a unique dynamic encoder that better leverages the memory graph features, as described below.


First, to improve the feature expression ability of the navigation thinking network, we use fully connected (FC) layers to map the features $ \widetilde{M}$ to higher dimensional spaces.
Inspired by some advanced works \cite{dosovitskiy2020image, liu2021swin} on vision transformers, we add layer normalization between the two FC layers to stabilize the forward input distribution and backpropagation gradient \cite{xu2019understanding}. The encoding details can be formulated as follows:
\begin{equation}
Y=\delta (LN(\widetilde{M}W^{M_1})W^{M_2})
\end{equation}
where $\delta$ denotes the ReLU function, $LN$ denotes layer normalization, and $W^{M_1} \in \mathbb{R}^{11 \times 16}$ and $ W^{M_2} \in \mathbb{R}^{16 \times 32}$ are learnable parameters.

Then, a multi-scale dynamic kernel is calculated to refine the target orientation features into implicit nodes. As shown in Figure~\ref{TAMSA}, we use three temporal convolution networks (TCNs) with different dilation rates $d$ to generate three dynamic kernels with distinct scales. It is worth noting that the TCN with $d=0$ degenerates to an FC layer. In the early stages of the ``navigate to'' phase, the TOMG contains fewer valid nodes; thus, the boundary degradation caused by zero padding has a greater impact. Accordingly, inspired by \cite{zhang2022edgeformer}, we design the circle padding (CP) which fills the sequence edge with the features at the other end of the sequence (Figure~\ref{TAMSA}). The different scale kernels are added after multiplying by the learnable parameter $w_d$: 
\begin{equation}
H(l)=\sum_{d=0}^{2}w_d(\sum_{j \in \Psi }Y(l+j* d)* f_d(j)+b_d)
\end{equation}
where $H=\{H(1),\cdots, H(L)\}$, $l$ is the central node of the convolution kernel, $\Psi$ refers to the set of offsets in the neighborhood considering convolution conducted on the central node, $Y(\cdot)$ takes out the node features in $Y$, and $f_d$ and $b_d$ denote the weights and biases in the convolution kernel with dilation rate $d$. The multi-scale dynamic kernel $H\in \mathbb{R}^{L \times \pounds}$ refines $Y\in \mathbb{R}^{L \times 32}$ to $\widetilde{Y}\in \mathbb{R}^{\pounds \times 32}$. 

Intuitively, the mappings between the observation data and target azimuth differ when searching for different targets. For example, when looking for a TV, even if the TV is located far from the agent, the agent can clearly identify the target and obtain a larger target bounding box; however, when looking for a mobile phone, the agent obtains a smaller target bounding box, even if the agent is close to the mobile phone. Therefore, we enhance the TAMSA representation by considering the target semantic information. To achieve this goal, the one-hot target index $E$ is encoded to the same channel dimension as $\widetilde{Y}$ through two FC layers, whose result is channel-wise multiplied with $\widetilde{Y}$ to get the target-aware feature representation $\widehat{Y}$:
\begin{equation}
\widehat{Y}=H^TY\odot \delta (\delta (EW^{E_1})W^{E_2})
\end{equation}

Finally, to obtain the final output $NT$ of the navigation thinking network, we flatten $\widehat{Y}$ from $\mathbb{R}^{\pounds \times 32}$ to $\mathbb{R}^{1 \times 32\pounds}$ and use an FC layer to reduce the output dimension. Furthermore, we add residual connections to ensure the stability of the feature transfer process.
\begin{equation}
NT=\delta (Flatten(\widehat{Y})W^Y)+\frac{1}{\pounds }\sum_{l=1}^{\pounds}\widehat{Y}(l)
\end{equation}
A dropout layer is added before the output to reduce  overfitting in the navigation thinking network.

\subsection{Adaptive Fusion (AF) of Dual Thinking Networks}
Search thinking and navigation thinking have different work strategies according to the navigation progress. During the “search for" phase, since there are no visited target-visible nodes, $NT$ is an all-zero matrix. Therefore, the navigation thinking network does not affect the action decision when the target has not yet been seen. During the “navigate to" phase, to ensure navigation robustness, search thinking and navigation thinking work together to guide the action decision. As the number of visited target-visible nodes increases, navigation thinking gradually dominates. The fusion process of the two thinking methods can be expressed as: 
\begin{equation}
DT=(LN(Concat(NT,ST)))W
\end{equation}
where $W$ is a learnable parameter matrix that adaptively adjusts the proportion of the two thinking networks, and LN is demonstrated to be significantly beneficial to the generalizability of the model.

\subsection{Policy Learning}

Following the previous works \cite{mirowski2016learning, fang2021target}, we treat this task as a reinforcement learning problem and utilize the asynchronous advantage actor-critic (A3C) algorithm \cite{mnih2016asynchronous}. However, in the search thinking network, the complex multi-head attention calculations are difficult to directly learn by the reinforcement learning \cite{du2021vtnet}; thus, we use imitation learning to pretrain the search thinking network. We divide the continuous action process into step-by-step action predictions and teach the agent to rely on only object associations to determine actions without considering historical navigation information. After pretraining, we obtain a search thinking network with a basic object association ability. Finally, the search thinking network and navigation thinking network are jointly trained via reinforcement learning.

\section{Experiment}
\subsection{Experimental Setup}
\subsubsection{Dataset}
AI2-Thor \cite{Kolve2017AI2THORAn} is our main experimental platform, which includes 30 different floorplans for each of 4 room layouts: kitchen, living room, bedroom, and bathroom. For each scene type, we use 20 rooms for training, 5 rooms for validation, and 5 rooms for testing. 

\subsubsection{Evaluation Metrics}
We use the success rate (SR), success weighted by path length (SPL) \cite{anderson2018evaluation}, and our proposed success weighted by navigation efficiency (SNE) metrics to evaluate our method. SR is formulated as $SR = \frac{1}{F}\sum_{i=1}^{F}Suc_i$, where $F$ is the number of episodes and $Suc_i$ indicates whether the $i$-th episode succeeds. SPL considers the path length more comprehensively and is defined as $SPL = \frac{1}{F}\sum_{i=1}^{F}Suc_{i}\frac{ \mathbb{L}_i^*}{max(\mathbb{L}_i, \mathbb{L}_i^*)}$, where $\mathbb{L}_i$ is the path length taken by the agent and $\mathbb{L}_i^*$ is the theoretical shortest path. SNE considers the navigation efficiency during the "navigate to" phase and is defined as
\begin{equation}
SNE=\frac{1}{F}\sum_{i=1}^{F}Suc_i\frac{\mathbb{L}_i}{\mathbb{L}_i^{nav}+1}
\end{equation}
where $\mathbb{L}_i^{nav}$ is the path length in the "navigate to" phase. To ensure that the denominator is nonzero, we use $\mathbb{L}_i^{nav} +1$ as the denominator in the above equation.

\subsubsection{Implementation Details}
We train our model with 18 workers on 2 RTX 2080Ti Nvidia GPUs. The dropout rate and target-visible filter $cf$ in our model are set to 0.3 and 0.4, respectively. The number of implicit nodes $\pounds$ in TAMSA is set to 3. We report the results for all targets (ALL) and for a subset of targets ($\mathbb{L}\ge 5$) with optimal trajectory lengths greater than 5.

\subsection{Ablation Experiments}

\begin{table*}[t]
\setlength\tabcolsep{5pt}
\caption{Ablation results on each module in the three sub-networks: search, navigate and fusion. }
\label{tab:overall ablation}
\begin{tabular}{c|ccccccc|ccc|ccc}
\hline
\multirow{2}{*}{ID} & \multicolumn{2}{c}{Search Thinking} & \multicolumn{3}{c}{Navigation Thinking} & \multicolumn{2}{c|}{Fusion} & \multicolumn{3}{c|}{ALL (\%)} & \multicolumn{3}{c}{$\mathbb{L}\ge 5$ (\%)} \\
&Associate    & Pretrain    & TOMG  & Egocentric  & TAMSA  & AF           & LN           & SR       & SPL     & SNE      & SR           & SPL          & SNE          \\ \hline \hline
 1&            &             &       &             &        &              &              & 71.34    & 43.47   & 121.91   & 60.72        & 42.18        & 110.73       \\ \hline
2& \checkmark          &             &       &             &        &              &              & 74.89    & 44.98   & 122.32   & 67.12        & 44.01        & 111.81       \\
3& \checkmark          & \checkmark         &       &             &        &              &              & 75.90    & 45.04   & 121.53   & 68.36        & 44.43        & 111.42       \\ \hline
4& \checkmark          & \checkmark         & \checkmark   &             &        &              &              & 76.02    & 43.15   & 126.15   & 68.66        & 42.19        & 119.22       \\
5& \checkmark          & \checkmark         & \checkmark   & \checkmark         &        &              &              & 78.12    & 42.01   & 129.12   & 70.52        & 41.23        & 121.87       \\
6& \checkmark          & \checkmark         & \checkmark   &             & \checkmark    &              &              & 78.04    & 45.67   & 131.24   & 70.34        & 45.30        & 125.98       \\
7& \checkmark          & \checkmark         & \checkmark   & \checkmark         & \checkmark    &              &              & 80.88    & 45.71   & 135.44   & 73.42        & 45.91        & 133.11       \\ \hline
8& \checkmark          & \checkmark         & \checkmark   & \checkmark         & \checkmark    & \checkmark          &              & 81.34    & 47.53   & 138.12   & 74.89        & 47.76        & 132.82       \\
9& \checkmark          & \checkmark         & \checkmark   & \checkmark         & \checkmark    & \checkmark          & \checkmark          & \textbf{82.39}    & \textbf{48.93}   & \textbf{139.83}   & \textbf{76.21}        & \textbf{49.32}        & \textbf{138.29}       \\ \hline
\end{tabular}
\end{table*}

\begin{table}[t]
\footnotesize
\setlength\tabcolsep{2.1pt}
\caption{Ablation experiments on each module in the target-aware multi-scale aggregator (TAMSA). Dynamic: dynamic aggregator kernel, TA: target-aware, MS: multi-scale, CP: circle padding.}
\label{tab:TAMSA}
\begin{tabular}{ccc|ccc|ccc}
\hline
\multicolumn{3}{c|}{\multirow{2}{*}{Method}}               & \multicolumn{3}{c|}{ALL (\%)} & \multicolumn{3}{c}{$\mathbb{L}\ge 5$ (\%)} \\
\multicolumn{3}{c|}{}                                      & SR      & SPL     & SNE      & SR           & SPL         & SNE          \\ \hline \hline
\multicolumn{3}{c|}{Average Pooling}                       & 79.67   & 45.14   & 134.21   & 73.33        & 45.21       & 126.94       \\
\multicolumn{3}{c|}{Transformer}                           & 77.23   & 43.24   & 132.83   & 71.44        & 42.97       & 127.11       \\
\multicolumn{3}{c|}{TCN}                                   & 78.66   & 43.41   & 133.69   & 74.42        & 43.91       & 125.18       \\ \hline
\multicolumn{1}{c|}{\multirow{4}{*}{TAMSA}} & A1 & Dynamic & 80.15   & 44.26   & 135.02   & 71.80        & 45.33       & 130.87       \\
\multicolumn{1}{c|}{}                       & A2 & A1+TA   & 81.20   & 46.71   & 136.25   & 74.17        & 47.52       & 134.91       \\
\multicolumn{1}{c|}{}                       & A3 & A1+MS   & 81.14   & 47.28   & 135.22   & 73.44        & 48.31       & 136.49       \\
\multicolumn{1}{c|}{}                       & A4 & A2+MS   & 81.32   & 47.41   & 137.26   & 75.88        & \textbf{49.36}       & 138.12       \\ 
\multicolumn{1}{c|}{}                       & A5 & A4+CP   & \textbf{82.39}   & \textbf{48.93}   & \textbf{139.83}   & \textbf{76.21}        & 49.32       & \textbf{138.29}       \\\hline
\end{tabular}
\end{table}

\subsubsection{Baseline}
Similar to \cite{dang2022unbiased}, our baseline model adopts the features concatenated from the image branch (from ResNet18), object branch (from DETR) and previous action branch as the environment perception encoding. Next, an LSTM network is used to model the temporal implicit features. The first row in Table~\ref{tab:overall ablation} shows the performance of our baseline. It is worth noting that since we adopt the object features extracted by DETR, the capability of our baseline model is already close to some SOTA methods with Faster-RCNN object detector \cite{ren2015faster}. 

\subsubsection{Dual Adaptive Thinking}
The purpose of dual adaptive thinking is to dynamically use two distinct thinking methods to ensure that the agent performs well during each stage. As shown in Table~\ref{tab:overall ablation}, the model with search thinking outperforms the baseline with the gains of 4.56/7.64, 1.57/2.25 and -0.38/0.69 in SR, SPL and SNE (ALL/$\mathbb{L}\ge 5$, \%). The search thinking network enables the agent to quickly locate the object through object associations; however, because the relative position of the target is not estimated, the SPL metric is limited by redundant paths in the ``navigate to'' phase. Adaptively incorporating our proposed navigation thinking into the search thinking improves the SR, SPL and SNE by 6.49/7.85, 3.89/4.89 and 18.3/26.87 (ALL/$\mathbb{L}\ge 5$, \%). The results prove that the navigation thinking improves the agent's performance on various indicators by optimizing the path during the "navigate to" phase. Moreover, the last two rows in Table~\ref{tab:overall ablation} show that the fusion of the dual thinking networks considerably improves the final model effect.

\subsubsection{Navigation Thinking Network}
The navigation thinking network includes three key modules: the target-oriented memory graph (TOMG), the egocentric coordinate transformation module and the target-aware multi-scale aggregator (TAMSA). Rows 4 through 7 in Table~\ref{tab:overall ablation} show the ablation results on the three modules. The navigation thinking network without the TAMSA increases the SR and SNE by 2.22/2.16 and 7.59/10.45, but decreases the SPL by 3.03/3.2 (ALL/$\mathbb{L}\ge 5$, \%). TAMSA improves the SPL back by refining the introduction of navigation thinking.

\begin{figure*}[t]
\centering
\includegraphics[width=\textwidth]{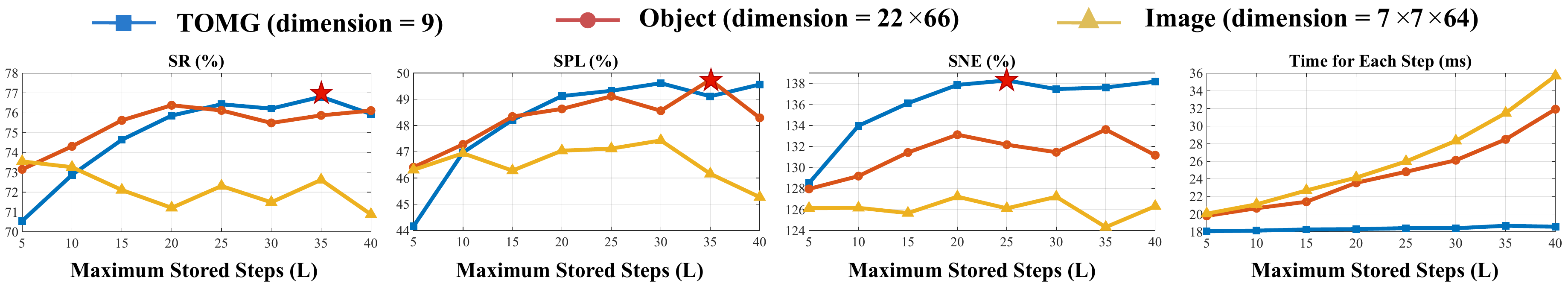} 
\caption{We compare the metrics in paths with $\mathbb{L}\ge 5$ while storing different features and path lengths for the navigation thinking. The red five-pointed star indicates the choices that optimize the given indicator.}
\label{TOMG}
\end{figure*}

Although the use of TOMG alone does not directly improve the performance of the various indicators, the simplified and highly abstract storage features in the TOMG facilitate the subsequent feature refinement and thinking integration.  Figure~\ref{TOMG} displays various metrics and computation speeds while using different storage features (TOMG, object and image) and maximum stored steps $L$. Image features with the most redundant information perform the worst. Compared with object features, the target-oriented characteristic in the TOMG considerably improves the SNE. Most importantly, the TOMG is substantially less complex than other storage methods in terms of calculation and memory costs. In terms of computational efficiency, when the number of stored steps is set to 40, compared with storing object and image features, the TOMG improves the computational speed by 41.43\% and 47.69\%, respectively. In terms of memory usage, the TOMG requires only 0.64\% and 0.29\% of the memory required by the object and image features. Furthermore, as the number of stored steps increases, the computational burden of the TOMG storage method remains essentially constant.

\subsubsection{Target-Aware Multi-Scale Aggregator (TAMSA)}

In contrast to the commonly used encoders such as TCNs and transformers, our proposed TAMSA uses a dynamic kernel to achieve automatic sequence length reduction without applying global pooling at the end. As shown in Table~\ref{tab:TAMSA}, the use of either TCNs or transformers exhibits worse performance than using average pooling directly. The results indicate that these commonly used encoders are not suitable for our navigation thinking network. Based on the initial aggregator model (A1), the target-aware (TA) property brings improvements of 1.05/2.37, 2.45/2.19, 1.23/4.04, and the multi-scale (MS) property brings improvements of 0.99/1.64, 3.02/2.98, 0.20/5.62 in SR, SPL and SNE (ALL/$\mathbb{L}\ge 5$, \%). The two properties optimize the agent's route in the “navigate to" phase during long-distance navigation but have little effect on the route during short-distance navigation. To address this issue, we utilize circle padding (CP) to prevent serious information loss in limited target-visible nodes, thereby optimizing the path during short-distance navigation. 

\subsubsection{Fusion of Dual Thinking Modules}

When humans complete a task, multiple thinking approaches often cooperate with each other rather than operating independently. Therefore, how to effectively integrate the two separately designed thinking networks through a unified network is crucial. Rows 4 to 7 in Table~\ref{tab:overall ablation} show that after the navigation thinking network is added to the model, the SPL  improves less than the SNE. This gap suggests that although navigation thinking optimizes the ``navigate to'' phase, it has a negative impact on the ``search for'' phase. Our proposed adaptive fusion (AF) method solves the above problem and improves the SR and SPL metrics by 0.46/1.47 and 1.82/1.85 (ALL/$\mathbb{L} >= 5$, \%). Moreover, since the feature modal and encoding methods used by the search thinking and navigation thinking are completely different, directly concatenating the two thinking features can lead to backpropagation instability and considerable overfitting. Therefore, layer normalization (LN) is used after the two thinking features are concatenated, improving the SR, SPL and SNE by 1.05/1.32, 1.40/1.56 and 1.71/5.47, respectively (ALL/$\mathbb{L} >= 5$, \%).

\subsection{Comparisons with the State-of-the-Art Methods}

Our DAT method is compared with three categories of relevant SOTA methods, as shown in Table~\ref{tab:SOTA}. \textbf{ (\uppercase\expandafter{\romannumeral1}) Methods with search thinking.} These methods have lower SNE because they do not apply navigation thinking. Compared to the recently proposed DOA \cite{dang2022unbiased} method, our DAT method brings 8.07/8.33, 8.66/8.96 and 18.97/29.10 improvements in SR, SPL and SNE (ALL/$\mathbb{L}\ge 5$, \%). \textbf{(\uppercase\expandafter{\romannumeral2})  Methods with long-term memory.} These methods theoretically depend on historical information to model environments more clearly; however, methods such as OMT \cite{fukushima2022object} store overcomplicated features, increasing the difficulty of network learning. Therefore, the current memory modules do not exert their full strength. \textbf{(\uppercase\expandafter{\romannumeral3}) Modular methods based on semantic maps.} The strong interpretability of semantic maps enables agents to quickly navigate to the target location after seeing the target; thus, their “navigate to” phase efficiency (SNE) is higher. Nevertheless, these methods require considerable efforts to explore the environment, resulting in an inability to visually capture targets as quickly as search thinking methods. The SPL of the current state-of-the-art modular method PONI \cite{ramakrishnan2022poni} is 11.66/12.92 lower (ALL/$\mathbb{L}\ge 5$, \%) than that of our DAT method.

\begin{table}[t]
\small
\setlength\tabcolsep{2.5pt}
\caption{Comparison with SOTA methods on the AI2-Thor dataset. More experiments on the other datasets are in the supplementary material (Table~\ref{tab:RoboTHOR}). }
\label{tab:SOTA}
\begin{tabular}{c|c|ccc|ccc}
\hline
\multirow{2}{*}{ID} & \multirow{2}{*}{Method}        & \multicolumn{3}{c|}{ALL (\%)}                           & \multicolumn{3}{c}{$\mathbb{L}\ge 5$ (\%)}             \\
        &                       & SR                        & SPL      & SNE                  & SR                        & SPL      & SNE                 \\ \hline \hline
\multirow{6}{*}{\uppercase\expandafter{\romannumeral1}} &
Random                         & 4.12                      & 2.21     & 7.91                  & 0.21                      & 0.08        & 8.14              \\
 &SAVN \shortcite{wortsman2019learning}                          & 63.12                     & 37.81         & 102.44             & 52.01                     & 34.94        & 94.51             \\
 &ORG  \shortcite{du2020learning}                          & 67.32                     & 37.01         & 111.88             & 58.13                     & 35.90         & 101.29            \\
 &HOZ \shortcite{zhang2021hierarchical}                           & 68.53                     & 37.50        & 110.79              & 60.27                     & 36.61        & 106.37             \\
 &VTNet  \shortcite{du2021vtnet}                        & 72.24 & 44.57 & 115.99 & 63.19 & 43.84 & 109.80\\
 &DOA  \shortcite{dang2022unbiased}                        & 74.32                     & 40.27            & 120.86          & 67.88                     & 40.36            & 109.19         \\ \hline
\uppercase\expandafter{\romannumeral2} & OMT  \shortcite{fukushima2022object}                        & 71.13                     & 37.27        &  124.31              & 61.94                     & 38.19        &  117.98             \\
\hline
\multirow{2}{*}{\uppercase\expandafter{\romannumeral3}} & SSCNav \shortcite{liang2021sscnav}       & 77.14                     & 35.09       & 138.22               & 71.73                     & 34.33       & 136.87              \\
 &PONI \shortcite{ramakrishnan2022poni}       & 78.58                     & 37.27           & \textbf{141.17}           & 72.92                     & 36.40           & 137.26          \\
\hline
\uppercase\expandafter{\romannumeral4}&\textbf{Ours (DAT)} & \textbf{82.39} & \textbf{48.93} & 139.83 & \textbf{76.21} & \textbf{49.32} & \textbf{138.29} \\ \hline
\end{tabular}
\end{table}

\begin{figure}[t]
\centering
\includegraphics[width=\columnwidth]{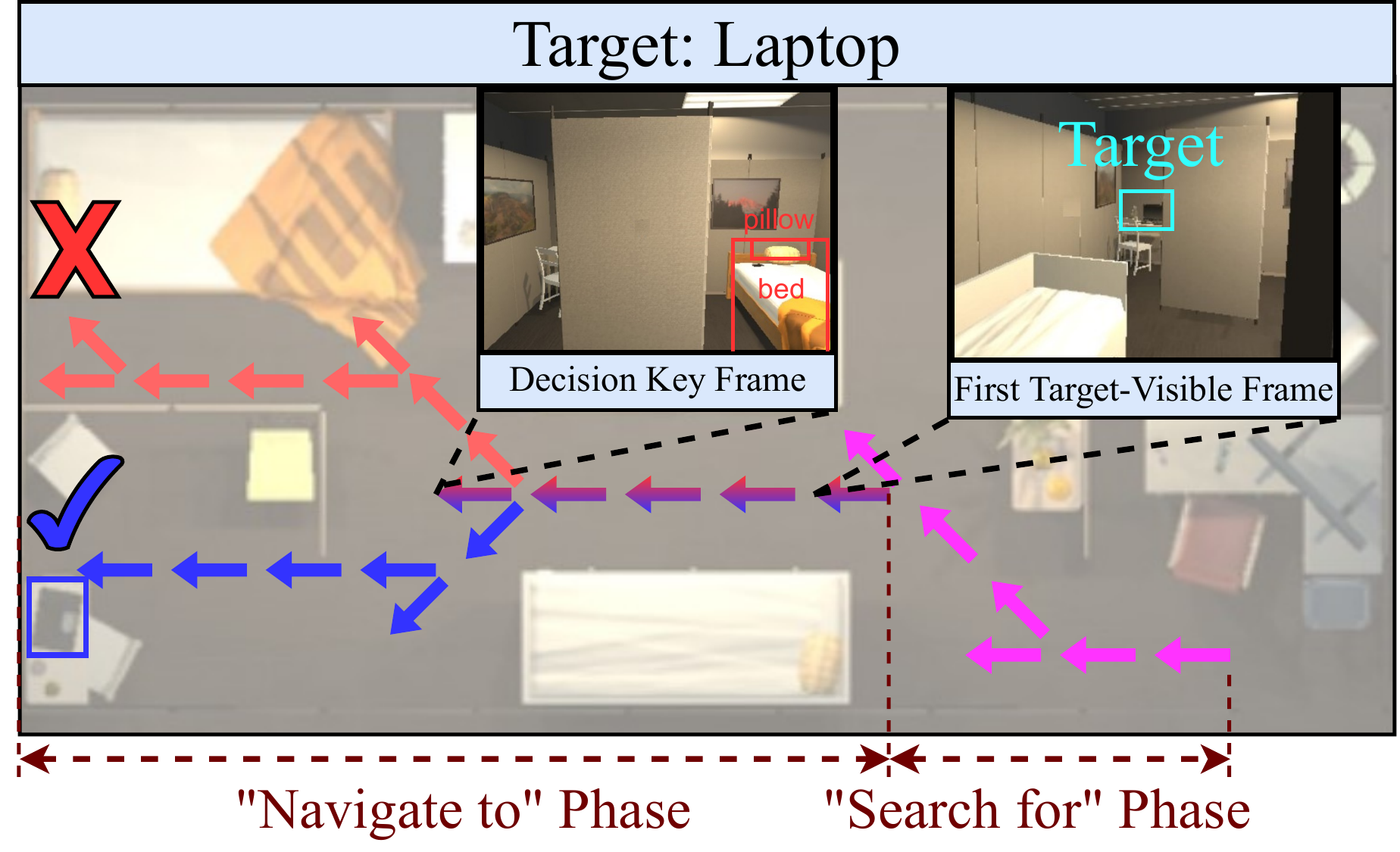} 
\caption{Visualization on the RoboTHOR test environment. Pink arrows: “search for" phase; red arrows: route with only the search thinking applied in the “navigate to" phase; blue arrows: route with both the search and navigation thinking applied in the “navigate to" phase. Both routes differ at the decision key frame.}
\label{visual_RoboTHOR}
\end{figure}

\subsection{Qualitative Analysis}

The route in the environment is visualized using only the search thinking and our DAT method, as shown in Figure~\ref{visual_RoboTHOR}. In the ``search for'' phase, the two methods predict the same path since our DAT method has not yet applied the proposed navigation thinking network. After entering the “navigate to" phase, the navigation thinking in the DAT method begins to assist the agent's decision-making by structuring the memory graph of the target information. At the decision key frame, the target cannot be seen. The method with only search thinking selects the right room with richer object information. In contrast, our DAT method uses the target relative position representation generated by the navigation thinking to select the correct left room. The correct decision in this key frame leads to the successful navigation of our DAT method. The route visualization in more scenarios is shown in the supplementary material (Figure~\ref{all_visual}).

\subsection{Multiple Adaptive Thinking in Embodied AI}

The dual adaptive thinking (DAT) network proposed in this paper provides key inspiration for future research. In the object navigation task, dual adaptive thinking can be extended to multiple adaptive thinking. The environment modeling thinking, object state understanding thinking, and other types of thinking can be introduced in multiple adaptive thinking models. Furthermore, multiple adaptive thinking is not limited to object navigation tasks. In other embodied AI tasks, such as embodied question answering (EQA) \cite{das2018embodied} and visual language navigation (VLN) \cite{anderson2018vision}, agents can use multiple thinking approaches to more flexibly address real-world problems.

\section{Conclusion}

In this paper, agents use a dual adaptive thinking (DAT) method to address the issue of not being able to quickly reach the target position after locating the target. Dual thinking includes the search thinking responsible for searching the target and the navigation thinking responsible for navigating to the target. The extensive experiments prove that dual adaptive thinking flexibly adjusts the thinking methods according to the navigation stage, thereby improving the success rate and navigation efficiency. It is worth noting that beyond the current object navigation task, multiple adaptive thinking can theoretically be applied to various time-series embodied AI tasks. 

\balance

\bibliography{aaai23}



\newpage
\appendix
\begin{table*}[htbp]
\centering
\setlength\tabcolsep{5pt}
\caption{The performance of methods with different thinking levels in the processes of “navigate to" and “search for".}
\label{tab:single-dual-human}
\begin{tabular}{c|ccc|cc|cc}
\hline
\multirow{2}{*}{Method} & \multicolumn{3}{c|}{Path Length} & \multicolumn{2}{c|}{Success Rate} & \multicolumn{2}{c}{Rotation Action Rate} \\
                        & ALL      & Search     & Navigate    & ALL           & Seen Target       & Search             & Navigate            \\ \hline \hline
Single Thinking         & $19.92_{\pm0.31}$    & $7.42_{\pm0.14}$    & $12.50_{\pm0.21}$     & $75.90\%_{\pm0.52\%}$       & $77.31\%_{\pm1.03\%}$           & $66.33\%_{\pm0.75\%}$            & $60.53\%_{\pm0.92\%}$             \\
Dual Thinking           & $18.26_{\pm0.20}$    & $7.96_{\pm0.11}$    & $10.30_{\pm0.18}$     & $82.39\%_{\pm1.21\%}$       & $92.71\%_{\pm0.82\%}$           & $71.62\%_{\pm0.79\%}$            & $49.04\%_{\pm0.66\%}$             \\
Human                   & $13.15_{\pm0.27}$    & $7.13_{\pm0.20}$    & $6.02_{\pm0.15}$     & $97.14\%_{\pm0.84\%}$       & $99.92\%_{\pm0.03\%}$           & $78.88\%_{\pm0.42\%}$            & $31.45\%_{\pm0.34\%}$             \\ \hline
\end{tabular}
\end{table*}

\begin{figure*}[hb]
\centering
\includegraphics[width=\textwidth]{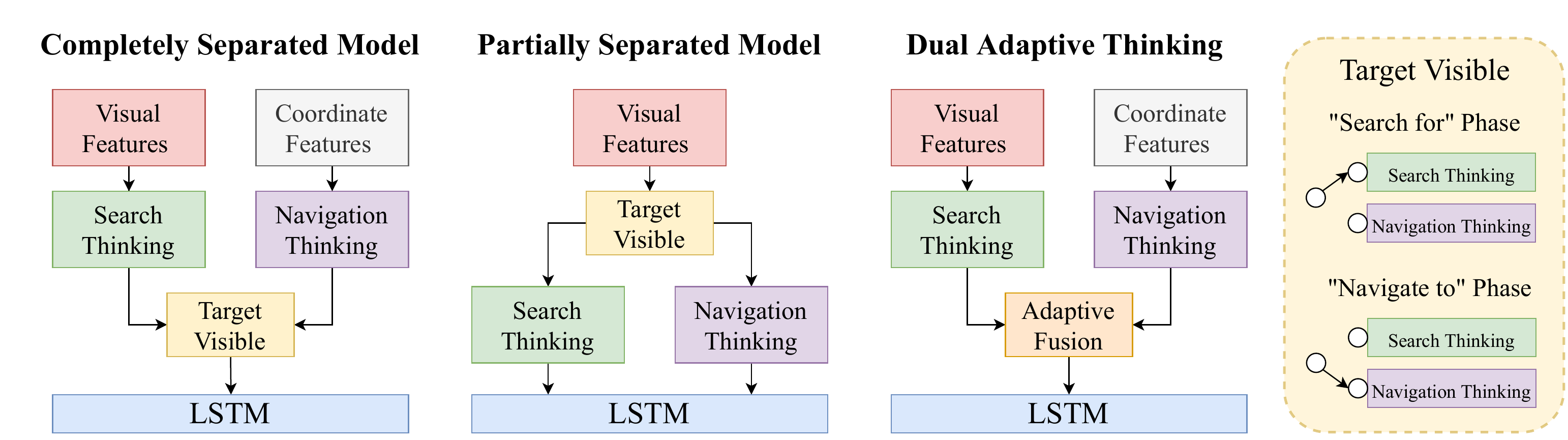} 
\caption{Different model structures for implementing dual thinking.}
\label{separate_methods}
\end{figure*}

\section{Disadvantages of Single Thinking}

In the introduction of the main text, we have explained the problems with single thinking in the “navigate to” phase. In this section, quantitative experimental results (Table~\ref{tab:single-dual-human}) will be used to explain in detail the navigation behavior gap between single thinking and human thinking. Moreover, how our dual adaptive thinking (DAT) approach bridges this gap will be presented.

In terms of the navigation path length, the total path length of the human result is 6.77 steps shorter than that of the single thinking method, which can be mainly attributed to path optimization in the “navigate to" phase. By introducing dual thinking, our DAT method significantly optimizes the route in the “navigate to" phase, but the navigation efficiency in the “search for" phase is slightly weakened. This problem is caused by the coupling of the two types of thinking in the joint training process, and we will try to solve this problem in future work.

Regarding the navigation success rate, from single thinking to dual thinking to human thinking, the success rate progressively increases. Notably, compared with the total success rate gap, if the agent has seen the target, the success rate gap between dual thinking and single thinking is much greater. The main reason is that the dual thinking method can make use of the navigation thinking network to help memorize and extract the relative target position features after seeing the target, thus guiding the decision for each step. Due to the limited environmental complexity of the dataset, the difference between the human's overall success rate and the success rate after seeing the target is not obvious. However, as the environmental complexity continues to increase, the gap between these two success rates of humans will become increasingly obvious.

Concerning the rotation and straight actions, as the thinking complexity increases, rotation actions are gradually concentrated in the “search for" phase. This trend is consistent with the different demand characteristics between the “search for" phase and the “navigate to" phase. In the “search for" phase, the agent needs to efficiently obtain environmental information and capture the target through abundant rotation actions. In the “navigate to" phase, the agent already knows the position of the target and needs to move forward to approach the target. The clear division of labor in each stage provides a guarantee of efficient navigation. 

\section{More Experiments}

\subsection{Different Dual Thinking Structures}

\begin{table*}[ht]
\centering
\setlength\tabcolsep{5pt}
\caption{Comparison with different model structures for implementing dual thinking.}
\label{tab:structures}
\begin{tabular}{c|ccc|ccc|c}
\hline
\multirow{2}{*}{Method} & \multicolumn{3}{c|}{ALL (\%)} & \multicolumn{3}{c|}{$\mathbb{L}\ge 5$ (\%)} & \multirow{2}{*}{Episode Length} \\
                        & SR       & SPL     & SNE      & SR           & SPL          & SNE           &                                 \\ \hline \hline
Partially Separated     & $77.49_{\pm0.56}$    & $46.61_{\pm0.44}$   & $125.57_{\pm1.47}$   & $69.48_{\pm0.82}$        & $45.62_{\pm0.61}$        & $116.94_{\pm1.52}$        & $27.19_{\pm0.36}$                           \\
Completely Separated    & $72.09_{\pm0.82}$    & $40.16_{\pm0.82}$   & $118.93_{\pm2.76}$   & $63.14_{\pm1.06}$        & $38.57_{\pm0.66}$        & $109.47_{\pm1.71}$        & $23.65_{\pm0.44}$                           \\
DAT (Ours)              & $82.39_{\pm0.91}$    & $48.93_{\pm0.68}$   & $139.83_{\pm2.01}$   & $76.21_{\pm1.11}$        & $49.32_{\pm0.83}$        & $138.29_{\pm1.77}$        & $18.26_{\pm0.36}$                           \\ \hline
\end{tabular}
\end{table*}

The essence of dual thinking is to use different decision networks in different navigation stages, which can be accomplished by means of various model structures. As shown in Figure~\ref{separate_methods}, in addition to our dual adaptive thinking (DAT) structure, we also experiment with completely and partially separated model structures.

The completely separated structure entirely decouples the thinking used in different stages. The difference with respect to our DAT method is that in the “navigate to" phase, the agent does not engage in search thinking and relies only on navigation thinking to make decisions. From Table~\ref{tab:structures}, it can be found that the navigation success rate of the completely separated structure is very low. There are two reasons for this model’s failure: (1) At the beginning of the “navigate to" phase, the target information in the target-oriented memory graph (TOMG) is not sufficient to generate a complete target orientation, so it cannot independently support action decisions. (2) If no search thinking is performed, visual information will be lost, resulting in a loss of obstacle avoidance ability. Therefore, in our DAT model, search thinking is active at all times. 

In the partially separated structure, search thinking and navigation thinking take the same input features but use different encoding networks to model the different types of thinking. The difference with respect to our DAT method is that instead of using our target-oriented coordinate features, navigation thinking relies on the same visual features as search thinking. As seen in 
Table~\ref{tab:structures}, the performance of the partially separated structure is not ideal, especially the episode length is too long. The comparison results suggest that the presence of too much redundant information in the visual features makes it too difficult for the navigation thinking network to learn the relative position of the target. More seriously, unreasonable navigation thinking affects the learning of the backbone visual embedding network, causing the whole model to collapse. 

Our DAT method uses specific input features and a reasonable adaptive fusion method for dual thinking to ensure stable and excellent network performance in different navigation stages.

\begin{table*}[t]
\centering
\small
\setlength\tabcolsep{5pt}
\caption{Comparison with SOTA methods on RoboTHOR \cite{deitke2020robothor}.}
\label{tab:RoboTHOR}
\begin{tabular}{c|ccc|ccc|c}
\hline
\multirow{2}{*}{Method} & \multicolumn{3}{c|}{ALL (\%)} & \multicolumn{3}{c|}{$\mathbb{L}\ge 5$ (\%)} & \multirow{2}{*}{Episode Length} \\
                        & SR       & SPL      & SNE     & SR            & SPL          & SNE          &                                 \\ \hline \hline
Random                  & $0.00_{\pm0.00}$     & $0.00_{\pm0.00}$     & $0.00_{\pm0.00}$    & $0.00_{\pm0.00}$          & $0.00_{\pm0.00}$         & $0.00_{\pm0.00}$         & $3.01_{\pm0.32}$                            \\
SP \cite{yang2018visual}                     & $27.43_{\pm0.60}$    & $17.49_{\pm0.36}$    & $42.27_{\pm0.87}$   & $20.98_{\pm0.97}$         & $16.03_{\pm0.48}$        & $32.78_{\pm0.79}$        & $68.18_{\pm1.27}$                           \\
SAVN \cite{wortsman2019learning}                   & $28.97_{\pm0.84}$    & $16.59_{\pm0.65}$    & $44.91_{\pm0.80}$   & $22.89_{\pm0.98}$         & $15.21_{\pm0.41}$        & $37.83_{\pm0.83}$        & $67.22_{\pm0.93}$                           \\
ORG   \cite{du2020learning}                  & $30.51_{\pm0.21}$    & $18.62_{\pm0.19}$    & $47.84_{\pm0.49}$   & $23.89_{\pm0.55}$         & $14.91_{\pm0.38}$        & $38.23_{\pm0.77}$        & $69.17_{\pm2.13}$                           \\
HOZ \cite{zhang2021hierarchical}                     & $31.67_{\pm0.87}$    & $19.02_{\pm0.49}$    & $50.42_{\pm0.97}$   & $24.32_{\pm0.48}$         & $14.81_{\pm0.23}$        & $41.10_{\pm0.79}$        & $66.26_{\pm1.88}$                           \\
VTNet   \cite{du2021vtnet}                 & $33.92_{\pm0.74}$    & $23.88_{\pm0.41}$    & $54.13_{\pm1.05}$   & $26.77_{\pm0.44}$         & $19.80_{\pm0.61}$        & $47.22_{\pm0.79}$        & $60.27_{\pm1.48}$                           \\
DOA  \cite{dang2022unbiased}                   & $36.22_{\pm0.37}$    & $22.12_{\pm0.42}$    & $57.09_{\pm0.95}$   & $30.16_{\pm0.28}$         & $18.32_{\pm0.34}$        & $48.15_{\pm0.86}$        & $61.24_{\pm1.23}$                           \\ \hline
\textbf{Ours (DAT)}              & $\textbf{41.72}_{\pm0.59}$    & $\textbf{27.91}_{\pm0.67}$    & $\textbf{74.21}_{\pm1.24}$   & $\textbf{37.16}_{\pm0.61}$         & $\textbf{22.46}_{\pm0.38}$        & $\textbf{67.82}_{\pm1.17}$        & $\textbf{58.81}_{\pm0.84}$                           \\ \hline
\end{tabular}
\end{table*}

\subsection{Comparisons on the RoboTHOR Dataset}

In the main text, for convenience of comparison, we mainly use AI2-Thor as the platform for algorithm verification. However, the AI2-Thor dataset is relatively simple and limited and thus cannot fully reflect the generalization ability of our method. Therefore, we additionally use the RoboTHOR dataset, which contains more complex scenes and longer navigation paths, to validate our method. 

RoboTHOR\cite{deitke2020robothor} is a platform for developing artificial embodied agents in simulated environments and testing them both in simulation and in the real world. A total of 89 scenes are available as simulated environments, of which 14 test scenes have counterparts constructed in the physical world. Because the simulators used in RoboTHOR and AI2-Thor are similar, the object navigation model we designed based on AI2-Thor could be directly trained and tested on RoboTHOR. 

Table~\ref{tab:RoboTHOR} illustrates the navigation ability of our DAT method and other state-of-the-art (SOTA) methods on RoboTHOR. Compared with the AI2-Thor dataset, the SR and SPL indicators both drop significantly on the RoboTHOR dataset, which indicates that the RoboTHOR dataset presents a greater challenge for object navigation algorithms. Nevertheless, our method still outperforms the SOTA methods by obvious margins of 5.50/7.00, 5.79/4.14 and 17.12/19.67 in terms of SR, SPL and SNE, respectively (ALL/$\mathbb{L}\ge 5$, \%). 

\begin{figure*}[t]
\centering
\includegraphics[width=\textwidth]{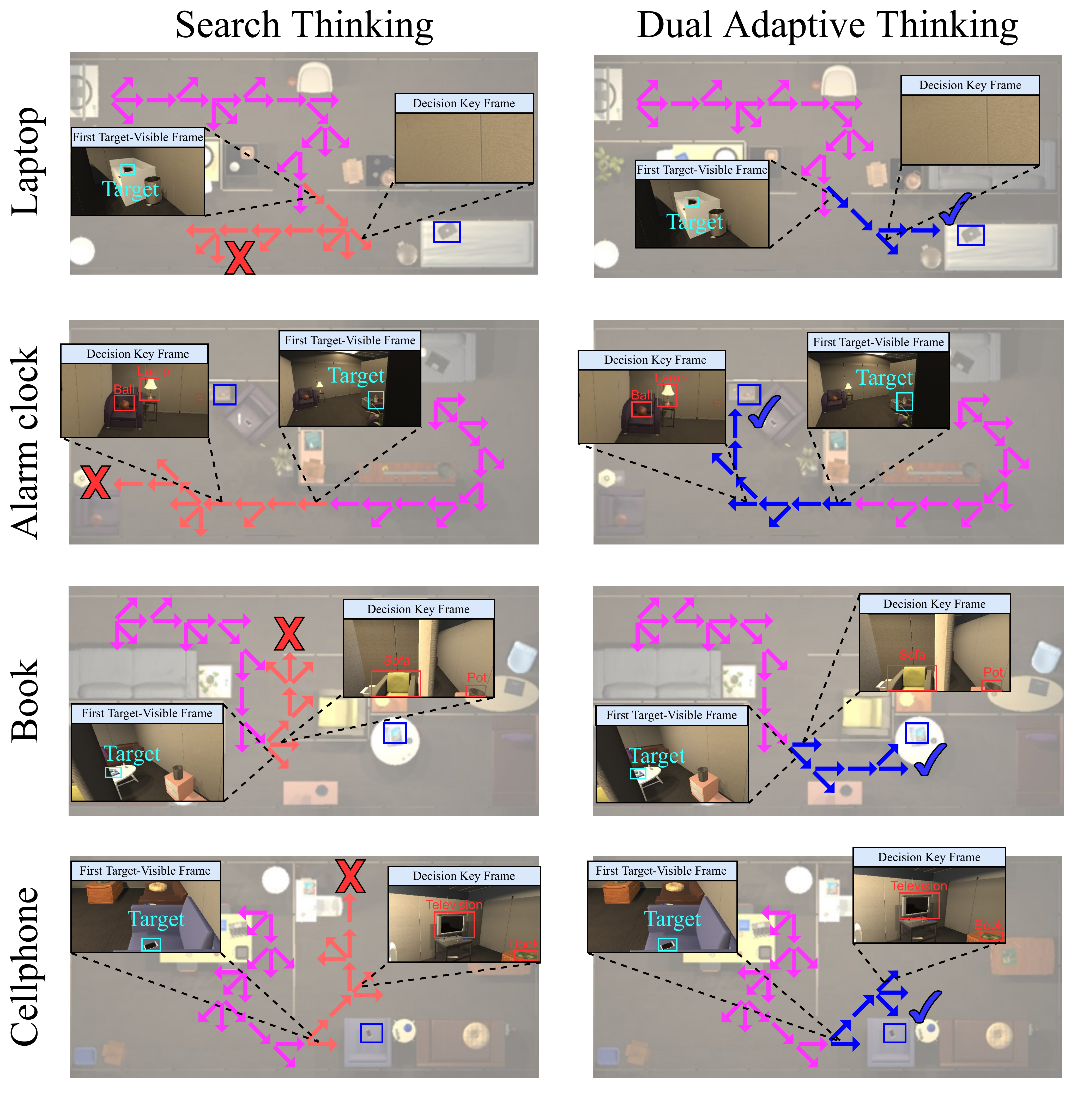} 
\caption{Visualization of navigation trajectories on the RoboTHOR dataset. The pink arrow indicates the path of the “search for" phase. The red arrow indicates the path of the “navigate to" phase when only search thinking is used. The blue arrow indicates the path of the “navigate to" phase when DAT is used.}
\label{all_visual}
\end{figure*}

\section{Qualitative Results}

In the main text, due to space limitations, we select only a simple scene for a qualitative analysis of navigation behavior. In Figure~\ref{all_visual}, we visualize the agent's navigation paths with different targets in four complex scenarios. We discover that when the agent needs to make a key decision, there is often no target in view. At these critical decision-making moments, our DAT method can provide the agent with the relative position information of the target, thereby improving the critical decision-making success rate. More significantly, our method makes each step of the agent more stable and purposeful in the “navigate to" phase. Reflecting on the navigation route, once the target has been seen, our method reduces the movement for in-situ exploration and applies more forward movement to navigate to the target position more efficiently.

\end{document}